\documentclass[lettersize,journal]{IEEEtran}
\usepackage{amsmath,amsfonts}
\usepackage[linesnumbered,ruled,vlined]{algorithm2e}
\usepackage{booktabs} 
\usepackage{array}
\usepackage[caption=false,font=normalsize,labelfont=sf,textfont=sf]{subfig}
\usepackage{textcomp}
\usepackage{stfloats}
\usepackage{url}
\usepackage{tabularx}
\usepackage{verbatim}
\usepackage{graphicx}
\usepackage{cite}

\usepackage{color}
\usepackage{xcolor}
\def\BibTeX{{\rm B\kern-.05em{\sc i\kern-.025em b}\kern-.08em
    T\kern-.1667em\lower.7ex\hbox{E}\kern-.125emX}}
\usepackage[pagebackref=true,breaklinks=true,letterpaper=true,colorlinks,citecolor=black, linkcolor=black,bookmarks=false]{hyperref}

\definecolor{darkyellow}{RGB}{255, 200, 0}
\begin{document}
\bibliographystyle{ieeetr}
\title{EduPlanner: LLM-Based Multi-Agent Systems for Customized and Intelligent Instructional Design}

\author{Xueqiao~Zhang\dag,
        Chao~Zhang\dag,
        Jianwen~Sun,
        Jun~Xiao,
        Yi~Yang,
        and~Yawei~Luo\ddag

\thanks{\dag~Equal contribution.}
\thanks{\ddag~Corresponding author: Yawei~Luo (yaweiluo@zju.edu.cn).}
\thanks{This work was supported by the National Natural Science Foundation of China (62293554, U2336212), National Key R\&D Program of China (SQ2023AAA01005), Zhejiang Provincial Natural Science Foundation of China (LZ24F020002),  Ningbo Innovation “Yongjiang 2035” Key Research and Development Programme (2024Z292), and Young Elite Scientists Sponsorship Program by CAST(2023QNRC001).}
\thanks{X.~Zhang, C.~Zhang and Y.~Luo are with the School of Software Technology, Zhejiang University, Ningbo, 315048, China.}
\thanks{J.~Sun is with the Faculty of Artificial Intelligence in Education,
Central China Normal University, Wuhan, 430079, China}
\thanks{J.~Xiao and Y.~Yang are with the College of Computer Science and Technology, Zhejiang University, Hangzhou, 310007, China.}
}

\maketitle
\begin{abstract}
Large Language Models (LLMs) have significantly advanced smart education in the Artificial General Intelligence (AGI) era. A promising application lies in the automatic generalization of instructional design for curriculum and learning activities, focusing on two key aspects: (1) Customized Generation: generating niche-targeted teaching content based on students' varying learning abilities and states, and (2) Intelligent Optimization: iteratively optimizing content based on feedback from learning effectiveness or test scores. Currently, a single large LLM cannot effectively manage the entire process, posing a challenge for designing intelligent teaching plans. To address these issues, we developed EduPlanner, an LLM-based multi-agent system comprising an evaluator agent, an optimizer agent, and a question analyst, working in adversarial collaboration to generate customized and intelligent instructional design for curriculum and learning activities. Taking mathematics lessons as our example, EduPlanner employs a novel Skill-Tree structure to accurately model the background mathematics knowledge of student groups, personalizing instructional design for curriculum and learning activities according to students' knowledge levels and learning abilities. Additionally, we introduce the CIDDP, an LLM-based five-dimensional evaluation module encompassing \textbf{C}larity, \textbf{I}ntegrity, \textbf{D}epth, \textbf{P}racticality, and \textbf{P}ertinence, to comprehensively assess mathematics lesson plan quality and bootstrap intelligent optimization. Experiments conducted on the GSM8K and Algebra datasets demonstrate that EduPlanner excels in evaluating and optimizing instructional design for curriculum and learning activities. Ablation studies further validate the significance and effectiveness of each component within the framework. Our code is publicly available at \url{https://github.com/Zc0812/Edu_Planner}.

\end{abstract}

\begin{IEEEkeywords}
Large Language Models, Multiple Agents, Smart Education, Instructional Design, Intelligent Agent.
\end{IEEEkeywords}

\section{\textbf{INTRODUCTION}}

\begin{figure}[t]
\centering
\includegraphics[width=\linewidth]{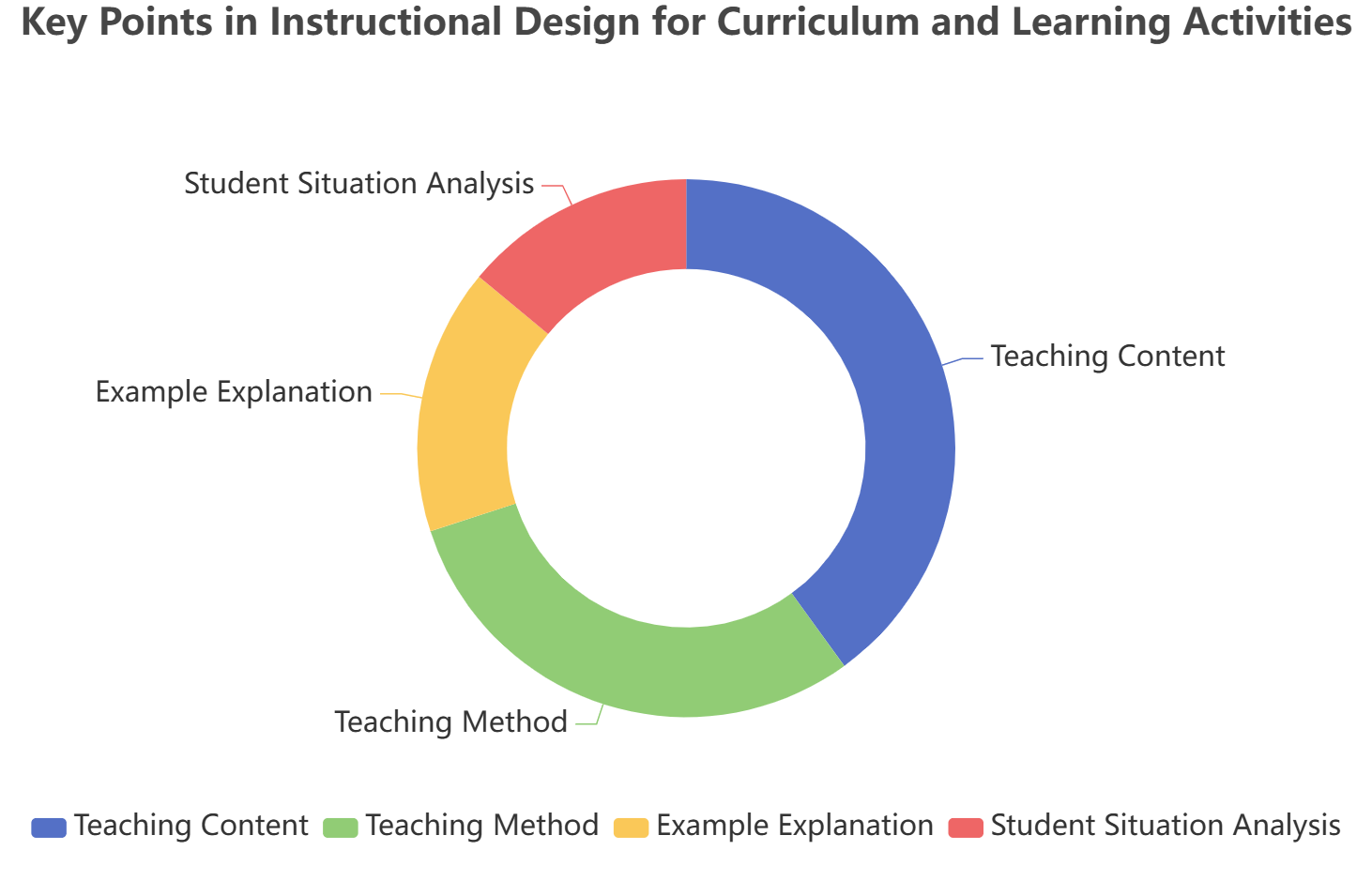}
\caption{The proportion of people who consider various factors critical in the task of generating lesson plans. Taking mathematics lessons as our example, in EduPlanner, we model the \textcolor{red}{student situation} using a newly proposed Skill-Tree ability module. We iteratively evaluate and optimize the \textcolor{green}{teaching method} and \textcolor{blue}{teaching content} through the adversarial collaboration between the Evaluator Agent and the Optimizer Agent. Additionally, an Analyst Agent is employed to analyze \textcolor{darkyellow}{error-prone examples} and incorporate them into the subsequent lesson plan. EduPlanner effectively addresses the most pressing challenges in lesson planning.}
\label{pie}
\end{figure}

\IEEEPARstart{I}{n} the educational realm, formulating pedagogical tasks presents a challenge that involves numerous scenarios and test cases. The crafting of instructional design for curriculum and learning activities~\cite{cevikbas2024empirical,liu2024automated} is an essential but difficult component of these tasks. It requires a considerable investment of time and energy from educators to understand the diverse learning profiles of their students before making an instructional design for curriculum and learning activities, ensuring a comprehensive structure, clarity of content, the inclusion of appropriate introductions and illustrative examples, as well as reflecting on and optimizing teaching behavior after class. To explore the pedagogical priorities that shape teachers' instructional designs, Wu \emph{et al.}~\cite{wu} revealed that $40.2\%$ educators consider teaching content the paramount aspect, $29.4\%$ place heightened emphasis on the application of teaching methods, and $16.3\%$ focus on example problem explanations. The remaining $14.1\%$ focus on an in-depth analysis of the student situation, as depicted in Fig.~\ref{pie}. Traditionally, teachers have relied on their accumulated experience and innate intuition to devise instructional frameworks. This process often necessitates a trial-and-error approach, with iterative improvements made through classroom implementation. Moreover, the diversity of student dispositions and the unpredictability of the classroom environment compound the difficulty of crafting comprehensive instructional design for curriculum and learning activities prior to instruction. The absence of a standard metric to evaluate the quality of these plans further exacerbates the complexity of this task, rendering the entire procedure laborious and often inefficient.

\IEEEpubidadjcol
The emerging trend in Generatine AI, distinguished by its ability to generate a diverse range of content, including texts, audio, images, and videos, has gradually penetrated the realm of education~\cite{tossell2024student,liu2024comet,liu2024scaffolding,ma2023personas,sun2023adversarial,zou2024balancing}, thus influencing various aspects of traditional pedagogy.
Recent advances in GAI, such as LLM, have positioned them as promising solutions for instructional design~\cite{xu2024eduagent,jinxin2023cgmi}. LLMs have the ability to develop sophisticated virtual environments and analyze complex social dynamics~\cite{chen2023agentverse,2023generative,qian2023communicative}. By observing agent responses and metric scores, researchers can gain insight into the effectiveness of various strategic approaches within instructional design. However, existing zero-shot instructional design generation methods~\cite{he2024evaluating}, which depend entirely on the capabilities of LLM, often simply follow instructions to perform tasks without addressing the individual needs of students. 
In addition, the existing methods~\cite{hu2024teaching} only concentrate on the generation and evaluation of instructional designs, yet they fail to incorporate the application of LLM in the optimization process of instructional designs.
Consequently, these instructional designs tend to be dull in knowledge transmission, deficient in case explanations, and devoid of a setting in course difficulty levels.

To address these challenges, this paper introduces ``EduPlanner'', an automated multi-agent system designed for the generation, evaluation, and optimization of instructional design. Taking mathematics lessons as our example, ``EduPlanner'' enables teachers to employ a variety of mathematical instructional design drafts, simulate classroom scenarios, and make targeted adjustments to instructional strategies based on feedback. The system comprises an evaluator agent, an optimizer agent, and a question analyst, which collaborate adversarially to generate customized and intelligent instructional design for curriculum and learning activities.

``EduPlanner'' begins by assessing the overall learning situation of the class using an innovative Skill-Tree structure, which informs the creation of teaching resources that align with individual learning trajectories, thus improving instructional efficacy and student engagement. Based on personal evaluation of the class, ``EduPlanner'' generates an initial instructional design for curriculum and learning activities and forwards it to the evaluator agent. The evaluator agent employs a five-dimensional evaluation system comprising \textbf{C}larity, \textbf{I}ntegrity, \textbf{D}epth, \textbf{P}racticality, and \textbf{P}ertinence (CIDPP). Feedback from the evaluator is then used by the optimizer agent to improve the instructional design and by the analyst agent to record typical examples. This adversarial cycle mimics a real teacher's iterative process of instructional design for curriculum and learning activities.

\begin{figure*}[!htbp]
    \subfloat[]{
        \includegraphics[width=\columnwidth]{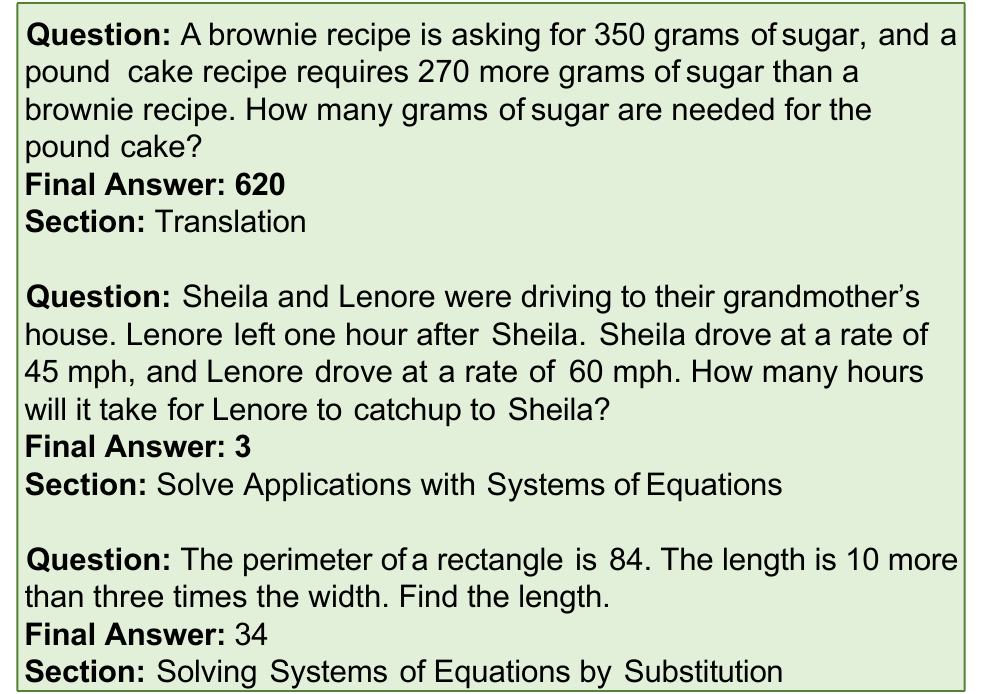}}
    \hfill
    \subfloat[]{
        \includegraphics[width=\columnwidth]{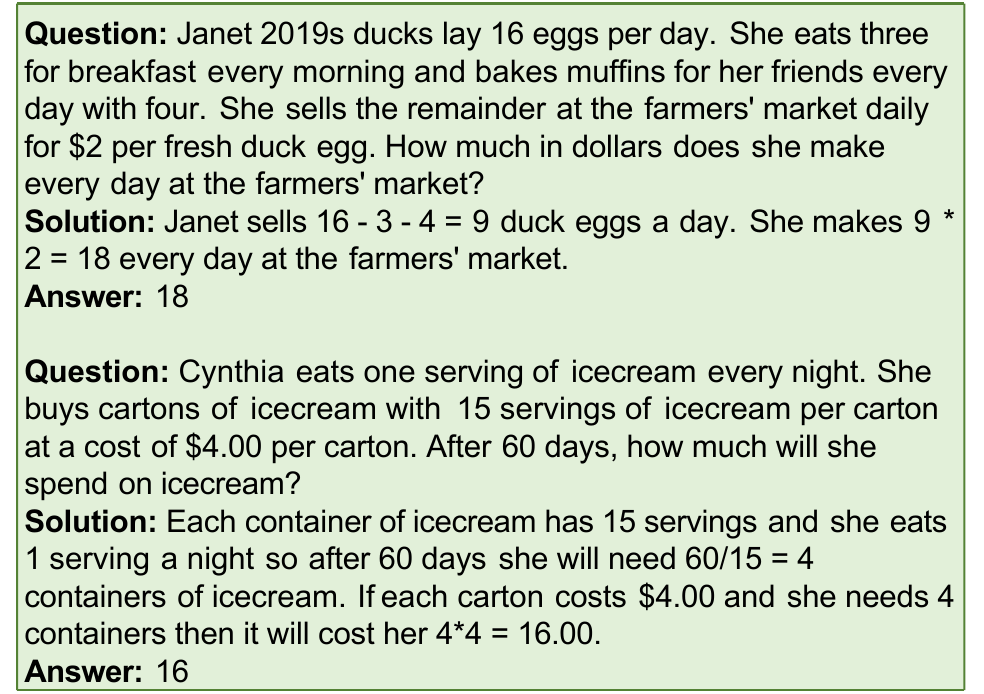}}
    \caption{Examples of datasets. On the left is GSM8K~\cite{cobbe2021training} and on the right is Algebra~\cite{he2023solving}. GSM8K is a dataset consisting of 8.5K high-quality, language-diversified elementary school mathematics word problems released by OpenAI. Algebra is a dataset comprised of 222 examples related to algebraic equations.}
    \label{dataset}
\end{figure*}

We evaluate the effectiveness of our framework using CIDPP-based experiments on the GSM8K~\cite{cobbe2021training} and Algebra~\cite{he2023solving} datasets, as illustrated in Fig.~\ref{dataset}. Compared to conventional methods, EduPlanner significantly improves the efficiency of instructional design for curriculum and learning activities, reduces pedagogical burdens, and facilitates precise alignment with the diverse educational needs of students.

The main contributions of this study can be summarized as follows: 
\begin{itemize}
\item{We introduce ``EduPlanner'', a novel framework for customized and intelligent instructional design including three agents: Evaluator Agent, Optimizer Agent, and Analyst Agent.}
\item{We propose an innovative Skill-Tree structure to craft a personalized learning profile for each student which provides corresponding input for instructional generators, enabling a focus on tailored student needs when structuring educational content.}
\item{We design a comprehensive five-dimensional instructional design evaluation system (CIDPP), and our framework achieves state-of-the-art results in experiments utilizing this system.}

\end{itemize}
\section{\textbf{RELATED WORK}}
\subsection{\textbf{Smart Education}}
Smart education, a modern campus feature, employs AI and technology to create an interactive learning environment and improve academic achievement~\cite{khan2024selection}.
Gao \emph{et al.}~\cite{gao2023understanding} employ AI for speech recognition and customized learning. 
Almogren \emph{et al.}~\cite{almogren2024exploring} use ChatGPT to provide instant assistance, encourage independent learning, and provide prompt feedback. 
Saurav Shrestha \emph{et al.}~\cite{shrestha2024developing} present the creation process of the Social Presence-Enabled Augmented Reality (SPEAR) tool, an innovative augmented reality (AR) learning application designed specifically for online engineering education.
Nicolas Pope \emph{et al.}~\cite{pope2025children} address the limit of educational tools to teach novice K-12 learners about data-driven systems by introducing the "GenAI Teachable Machine," a visual, data-driven design platform.
CV Suresh and Bhavesh \emph{et al.}~\cite{babu2025ai} explore the transformative potential of AI-driven instructional design to enhance personalized learning experiences and improve educational efficiency.

\subsection{\textbf{Generative AI for Educational Transformation}}
GAI has attracted great attention with the advancement of LLMs. 
Within the educational domain, GPTeach~\cite{markel2023gpteach} employs GPT to simulate student behaviors. EduAgent~\cite{xu2024eduagent} excels in anticipating the detailed learning actions of students and generating plausible learning behaviors for virtual students without ground truth. 
AgentVerse~\cite{chen2023agentverse} enables the development, training, and interaction of agents across various domains, simulating complex human decision-making processes. 
Gradescope~\cite{singh2017gradescope} improves the assessment process for educators by organizing assessment tasks, increasing student engagement, and providing valuable data on student performance.

\subsection{\textbf{LLMs-based Multi-agent Systems}}
Recent studies have explored the use of LLMs in agent systems, demonstrating excellent performance in accomplishing tasks~\cite{li2024agent,qian2023communicative} upon prompts through collaborative interactions among multiple agents. The concept of generative agents~\cite{park2023generative} involves the creation of a simulated town with $25$ agents. 
The approach relies on multiple language models~\cite{wang2023voyager,luo2024large}, often proves inefficient, particularly in open-ended settings.
Managing the control aspects of LLM agents, Zheng \emph{et al.}~\cite{zheng2023steve} presents a challenge, yet this is where reinforcement learning~\cite{lifshitz2024steve,krishna2022socially} excels.
MetaGPT~\cite{hong2023metagpt} uses standard operating procedures to enhance the multi-agent systems and emulates a software team with agents including product manager, architect, project manager, engineer, and quality engineer.
Autogen~\cite{wu2023autogen} offers designable conversable agents and is capable of integrating with external tools to facilitate collaborative conversations within multi-agent systems.
AutoGPT~\cite{yang2023auto} works based on established goals without human intervention and employs a real-time loop evaluation strategy to assess the completion of objectives dynamically.
CAMEL~\cite{li2023camel} develops a role-playing framework that generates dialogue datasets to foster communication and collaboration within multi-agent systems. 
\begin{figure*}[!htbp]
\centering
\includegraphics[width=\textwidth]{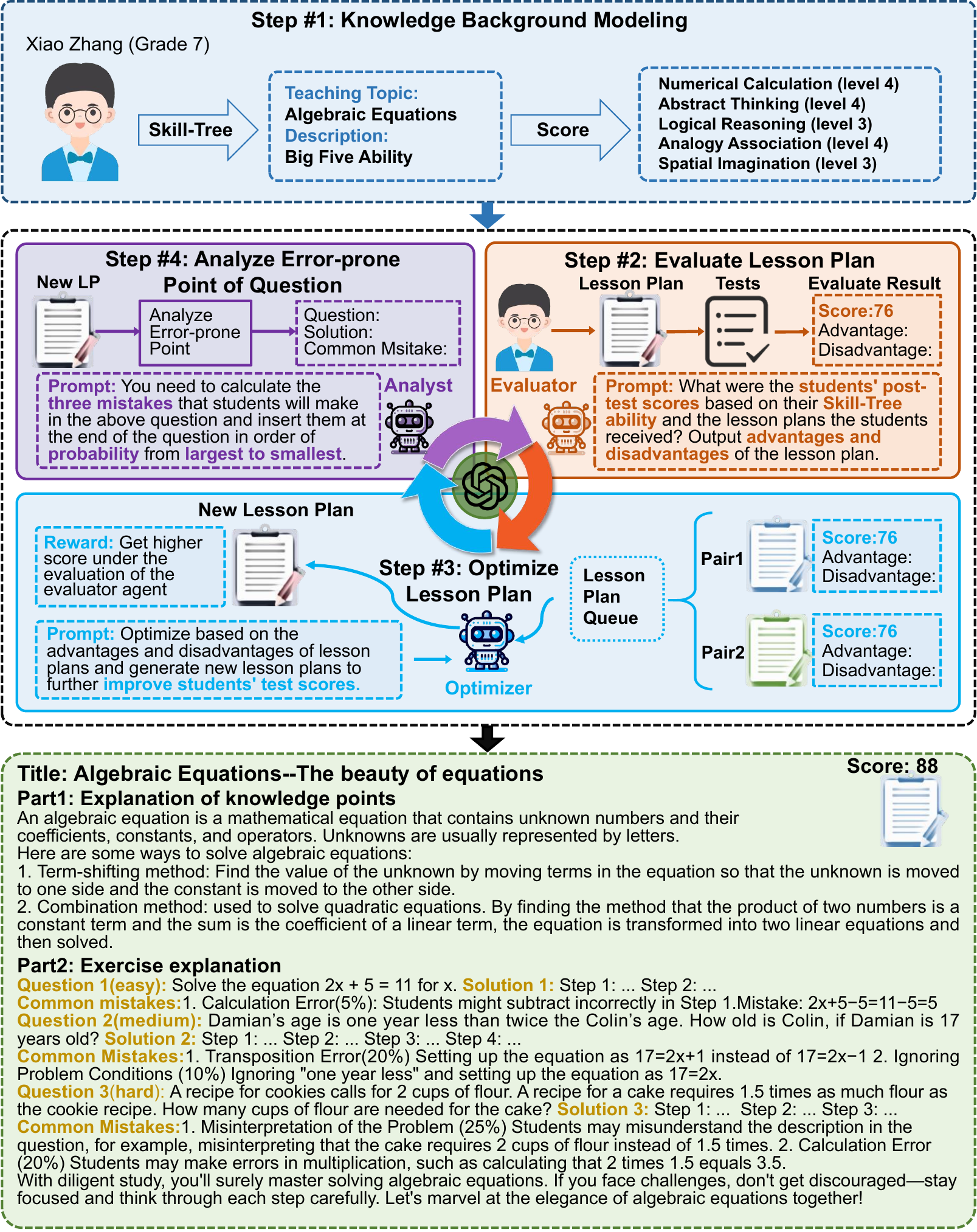}
\caption{{EduPlanner: LLM-Multi-Agent-Based Instructional Design Generation Framework. When the optimization is completed, the instructional design for curriculum and learning activities with the highest score in the queue is output, and its content mainly includes two parts: knowledge point explanation and question explanation.}}
\label{framework}
\end{figure*}
\subsection{\textbf{Evaluation using LLMs}}
Most of the evaluations in relative experiments are open-ended and lack definitive reference answers, designing a rule-based program to evaluate the output of LLM is challenging. Traditional evaluation metrics based on the similarity between the output and the reference answers (\emph{e.g.}, ROUGE~\cite{lin2004rouge}, BLEU~\cite{papineni2002bleu}) also fail to address these questions. Employing experts for manual evaluation is time-consuming and expensive. 
Therefore, in order to facilitate an efficient automated evaluation process, some works explore LLMs as a substitute for human evaluation. These models, often trained via Reinforcement Learning from Human Feedback (RLHF), have demonstrated great alignment with humans. 
The LMSys~\cite{zheng2024judging} explores the merits and limitations of utilizing diverse LLMs as judges for tasks ranging from writing and mathematics to general world knowledge.
The self-rewarding LLM~\cite{yuan2024self} trains itself through the data generated by its own and evaluated by the GPT-4. 
CharacterLLM~\cite{shao2023character} employs GPT as an evaluator to access the LLMs in role-playing scenarios, and Neeko~\cite{yu2024neeko} evaluates the performance on role switching of LLM in multi-role-playing contexts based on GPT. 
L-eval~\cite{an2023eval} takes the pair-wise battle format and reports the win rate \emph{vs.} LLMs, augmenting the judgment prompt to mitigate the preference for more detailed and lengthy answers. 

\section{\textbf{METHODOLOGY}}

\subsection{\textbf{Overview}}

We present EduPlanner, a multi-agent-based system designed for the evaluation and optimization of instructional designs. The system employs a modular architecture, depicted in Fig.~\ref{framework}. To tailor instructional design for specific student groups, we model students' knowledge backgrounds using a Skill-Tree structure. This structure facilitates the generation and optimization of instructional designs that meet the diverse learning needs of students. Our approach includes three collaborative agents: the Evaluator Agent, which assesses the quality of instructional designs; the Optimizer Agent, which improves lesson content; and the Analyst Agent, which identifies common error points in example questions within the instructional designs. These agents are integrated using an adversarial learning framework. The Evaluator Agent provides feedback on the advantages and disadvantages of the instructional designs, guiding the Optimizer Agent and Analyst Agent. The evaluation scores serve as targets for optimization, driving an iterative adversarial process that enhances the instructional design for curriculum and learning activities. When optimization is complete, the instructional design with the highest score in the queue is output and its content mainly includes two parts:  knowledge point explanation and question explanation.

\subsection{\textbf{Skill-Tree Structure}}
\begin{figure*}[!htbp]
\centering
\includegraphics[width=\textwidth]{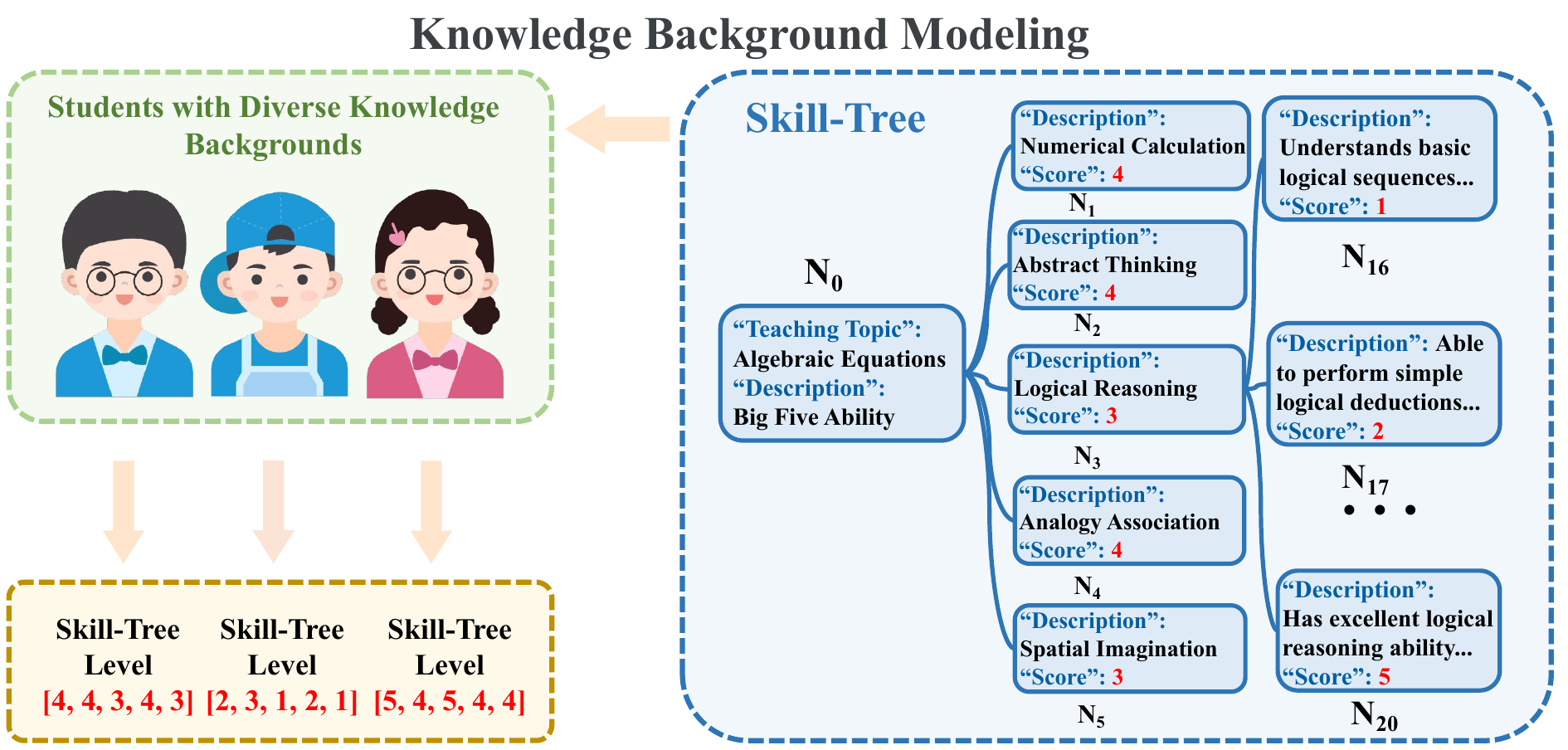}
\caption{Skill-Tree structure models the students' diverse knowledge background through five principal abilities, outputs the students' Skill-Tree level, and provides it to the agents so that the agents can capture the knowledge background of different students and ensure that targeted instructional design for curriculum and learning activities can be generated.}
\label{skill-tree}
\end{figure*}

To generate customized instructional design for students with different knowledge backgrounds, a key issue to be addressed is the modeling of students' abilities. In CGMI~\cite{jinxin2023cgmi}, a character tree structure was designed to solve the problem of inconsistency of roles in LLM after multiple rounds of dialogue.
Following CGMI~\cite{jinxin2023cgmi}, we choose to employ a tree structure to portray the competencies\footnote{We use mathematics as an example to illustrate the role of the Skill-Tree structure, while other subjects can also be applied.} of students. In a novel extension of this approach, we introduce five specialized sub-nodes under each primary node. This enhancement enables a detailed quantification of abilities across different dimensions, offering an in-depth description of the knowledge bases and learning abilities of individual students, as shown in Fig.~\ref{skill-tree}. Compared to traditional binary evaluations (good or bad), this approach offers a more comprehensive analysis with multiple dimensions, enabling the evaluator agent to capture the differences in students' abilities.

\subsubsection*{\bf Student Abilities (Mathematics as Example)}
\begin{itemize}
\item{Node N\textsubscript{0}: explains the content of the teaching topic and the description of nodes N\textsubscript{1}--N\textsubscript{5}}
\item{Node N\textsubscript{1}: represents Numerical Calculation ability, reflecting the accuracy and proficiency of students in addition, subtraction, multiplication, and division.}
\item{Node N\textsubscript{2}: represents Abstract Thinking ability, one of the key factors in understanding and solving algebraic equations.}
\item{Node N\textsubscript{3}: depicts Logical Reasoning ability, a core skill for solving mathematical problems.}
\item{Node N\textsubscript{4}: represents Analogy Association ability, an important tool for students to expand knowledge boundaries.}
\item{Node N\textsubscript{5}: refers to Spatial Imagination ability, which is particularly important for solving geometric problems in algebraic equations.}
\end{itemize}

\subsection{\textbf{Instructional Design Evaluation}}
In educational practice, the effectiveness of instructional design is a challenging issue. Given the great performance on various tasks of LLM-Based-Agents, we are encouraged to utilize agents as experienced evaluation experts that provide feedback on instructional designs, such as advantages, disadvantages, and suggestions for improvement.
\subsubsection{\textbf{Evaluation Criteria}}
 To comprehensively evaluate instructional designs and catalyze intelligent optimization, we introduce the CIDDP, an LLM-based five-dimensional evaluation module encompassing \textbf{C}larity, \textbf{I}ntegrity, \textbf{D}epth, \textbf{P}racticality, and \textbf{P}ertinence.
 This evaluation system comprises the pedagogical philosophy in instructional design, with each dimension serving a distinct purpose:
 
\textbf{C}larity is essential for making instructional designs clear and removing unnecessary information. It helps teachers establish clear teaching goals, ensuring that educational activities are aligned with pedagogical principles of purposeful instruction.

\textbf{I}ntegrity involves examining instructional designs to ensure they are detailed and complete, covering all necessary aspects like clear explanations of concepts and relevant examples. It aligns with the pedagogical philosophy that promotes a comprehensive and integrated approach to learning.

\textbf{D}epth measures how instructional designs engage students in deep thinking and present content that reveals the deep meanings and relationships within knowledge. It is in line with the pedagogical emphasis on fostering deep learning experiences.

\textbf{P}racticality assesses the real-world applicability of instructional designs and how well students can use their knowledge to solve practical problems. It corresponds with the pedagogical philosophy that focuses on applying learned knowledge.

\textbf{P}ertinence examines the efficacy of instructional designs in satisfying the diverse learning needs of students. It promotes the adaptation of different levels of knowledge and learning preferences, reflecting the personalized teaching approach central to modern pedagogical thought.

\subsubsection{\textbf{Evaluator Agent}}

The Evaluator Agent, built on the \texttt{Meta-Llama-3-70B} LLM~\cite{meta2024}, is designed to evaluate instructional designs and offer feedback. Trained on $100$ carefully annotated instructional design for curriculum and learning activities, each assessed by education experts for effectiveness, scoring, advantages, and disadvantages, the Evaluator Agent demonstrates an evaluation capability comparable to that of human experts. The process involves analyzing students' Skill-Tree scores to measure their knowledge levels and accessing scores after instruction of instructional designs using randomly selected test questions. The agent provides detailed feedback that clearly points out the advantages and content that needs improvement, forming a crucial basis for the subsequent optimization of the instructional design.

\subsubsection{\textbf{Instructiona Design Evaluation Algorithm}}
The Evaluator Agent assesses the quality of instructional design and provides feedback, denoted as $\mathbf{F}$ (\emph{e.g.}, advantages and disadvantages), to the Optimizer Agent. We refer to this algorithm as Evaluator Agent Expert Evaluation (EAEE)~\cite{he2024evaluating}. Specifically, we allow the Evaluator Agent understand students' background knowledge and learning abilities based on their Skill-Tree scores. Then, we provide the students with a specific instructional design for curriculum and learning activities and let the evaluator agent evaluate the students' average scores on exam questions. The specific mathematical Eq.~\ref{eq1} are as follows. The evaluator agent provides feedback on the instructional design for curriculum and learning activities. Alg.~\ref{Alg1} details the EAEE process.

\begin{algorithm}[t]
\label{Alg1}
\small
\caption{Evaluator Agent Expert Evaluation (EAEE)}
\KwIn{Evaluator Agent $\mathcal{A}_e$,
Initial Instruction Design $\mathcal{L}_{p_0}$, Knowledge Background, Evaluation Task, Test Questions, $\mathcal{X}_1$,$\mathcal{X}_2$,$\mathcal{...}$,$\mathcal{X}_T$}
\KwOut{post score $s$, advantage $a$, disadvantage $d$
}

Set knowledge background $\Vert$ Instruction Design $\Vert$ $\mathcal{X}_1$ → $\mathcal{P}_1$

Give $\mathcal{P}_1$ as input to ${\mathcal{A}_e}$ and receive output $\mathcal{O}_1$

\For{$i$ in $T$}{
Set $\mathcal{X}_i$ $\Vert$ evaluation task → $\mathcal{P}_i$

Give $\mathcal{P}_i$ as input to ${A_e}$ and receive output $\mathcal{O}_i$
}
Extract $s_1$,$\mathcal{...}$, $s_T$ from $\mathcal{O}_1$,$\mathcal{...}$, $\mathcal{O}_T$

Set ${\sum_{i}^{T} s_i / T}$   $\mathcal{→s}$

Extract $a_1$, $\mathcal{...}$, $a_T$, $d_1$, $\mathcal{...}$, $d_T$ from $\mathcal{O}_1$, $\mathcal{...}$, $\mathcal{O}_T$

Use ${A_e}$ to summarize $a_1$, $\mathcal{...}$, $a_T$  → $a$,  $d_1$, $\mathcal{...}$, $d_T$ → $d$

return $(s, a, d)$
\end{algorithm}

\begin{equation}
\label{eq1}
        R(lp,S) = \frac{1}{{|T|}}\sum\limits_{i = 1}^{|T|} f \left( {lp,S,{x_i}} \right),
\end{equation}
where $x_i$ represents the $i$-th test question, $lp$ is the instructional design to be evaluated, and $S$ is the Skill-Tree score of the student, allowing the LLMs to capture differences in students' learning abilities. The function $f(\cdot)$ is the evaluation function of the Evaluator Agent. $T$ is a hyperparameter we set, which represents the number of test questions used during an instructional design for evaluation. $R(\cdot)$ represents the average test score evaluated by the Evaluator Agent for students with test questions.

The process of the EAEE algorithm is as follows. The student role description, instructional design, evaluation task description, and the first test question are combined to form the initial prompt input $p_1$. The initial input $p_1$ is submitted to the Evaluator Agent, and its output $o_1$ is received, which includes predictions of the student's learning effectiveness. The process is repeated for the $M$ test questions. Finally, the model extracts the predicted probabilities or scores for the student's learning success from each output and calculates the average. The advantages and disadvantages of the instructional design generated in each output are also extracted and summarized by $A_e$. The summary is provided to the Optimizer Agent $A_o$ as reference information for optimization of the instructional design.

\subsection{\textbf{Instructional Design Optimization}}
\subsubsection{\textbf{Optimizer Agent}}
The Optimizer Agent is designed to optimize personalized, high-quality instructional designs according to the students' knowledge backgrounds. The process involves using the feedback $\mathbf{F}$ provided by the Evaluator Agent (\emph{i.e.}, the advantages and disadvantages of the instructional design) for optimization. The goal is to maximize the instructional design's evaluation score assessed by the Evaluator Agent. Eq.~\ref{eq2} expresses this instructional design optimization strategy.
\begin{equation}
    \label{eq2}
    \left\{ \begin{array}{l}
    F = f(lp,S)\\
    l{p_{new}} = \arg {\max _{R(l{p_{new}},S)}}{\rm{\{ g}}(D,F)\} 
    \end{array} \right.
\end{equation}

In this equation, $F$ represents the feedback generated by the Evaluator Agent, the function $g(\cdot)$ is responsible for optimizing and generating a new instructional design using the feedback and the instructional designs in the queue $D$ along with their scores, $R(\cdot)$ represents the evaluation score for the new instructional design. The final generated $lp_{new}$ is optimized to maximize the instructional design's evaluation score.

Our Optimizer Agent employs a dynamically sorted instructional design queue to prioritize the most effective instructional design. After the Evaluator Agent evaluates the new instructional design, which involves generating scores and feedback, the Optimizer Agent inserts the instructional design for the curriculum and learning activities into the queue based on its score. The process of the instructional design queue is described by Eq.~\ref{eq3}.

\begin{equation}
\small
    \label{eq3}
     {D_{n}} = {\mathop{\rm Trim}\nolimits} \left( {{\mathop{\rm Sort}\nolimits} \left( {{\mathop{\rm Ins}\nolimits} \left( {l{p_{{\rm{n}}}},D} \right),R(l{p_{n}},S)} \right),P} \right)
\end{equation}

In Eq.~\ref{eq3}, $D$ denotes the instructional design queue, $lp_{n}$ is the new instructional design, $S$ represents the Skill-Tree scores of the students, $P$ indicates the maximum capacity of the instructional design queue, $R(\cdot)$ represents the student's average test score, $\text{Ins}(\cdot)$ indicates inserting the new instructional design into the instructional design queue, $\text{Sort}(\cdot)$ means sorting the queue in ascending order, $\text{Trim}(\cdot)$ keeps only the top $P$ instructional design.

\subsubsection{\textbf{Instructional Design Optimization Algorithm}}

We refer to the instructional design optimization algorithm as Optimizer Agent Expert Optimization (OAEO)~\cite{he2024evaluating}. The agent optimizer ($A_o$) and the agent example analyzer ($A_a$) form the core of the optimization process. The student description, initial instructional design, evaluation task, optimization task, and the first test question are combined into the initial prompt ($p_0$). This prompt is submitted to the $A_o$, which generates an optimized instructional design for curriculum and learning activities output ($o_0$), including specific suggestions for content improvement.

The process is repeated for each test question. The $A_a$ extracts examples from each output, forming a set of examples. All suggestions and examples are combined into the final instructional design for curriculum and learning activities. Alg.~\ref{Alg2} provides a detailed description of the optimization process.

\begin{algorithm}[t]
\label{Alg2}
\small
\caption{Optimizer Agent Expert Optimization (OAEO)}
\KwIn{
    Evaluator Agent $\mathcal{A}_{\!E}$, 
    Optimizer Agent $\mathcal{A}_{\!O}$, 
    Question Analyst $\mathcal{A}_{\!A}$, 
    Initial Instructional Design $\mathcal{L}_{p_0}$, 
    Knowledge Background $\mathcal{K}$, 
    Evaluation Task, 
    Optimization Task, 
    Test Questions $\mathcal{X}_1, \mathcal{X}_2, \ldots, \mathcal{X}_T$,
    Optimization Rounds $\mathcal{N}$
}
\KwOut{
    Instructional Design $\mathcal{L}_{p_0}, \mathcal{L}_{p_1}, \ldots, \mathcal{L}_{p_{p-1}}$
}

Evaluate $\mathcal{L}_{p_0}$ using Algorithm 1 and receive score $\mathcal{R}_0$.

Set $\{(\mathcal{L}_{p_0}, \mathcal{R}_0)\} \to \mathcal{D}$.

Set knowledge background optimization task $\to \mathcal{P}_0$.

\For{$i$ in $\mathcal{N}$}{
    Set $[\,] \to D'$.
    
    \For{$k$ in $\mathcal{K}$}{
        Generate $\mathcal{C}_{n-1}^{\mathcal{K}+k}$ by feeding $\mathcal{P}_{n-1}$ into $\mathcal{A}_o$.
        
        Insert $q$ into $\mathcal{C}_{n-1}^{\mathcal{K}+k}$.
        
        Evaluate $\mathcal{C}_{n-1}^{\mathcal{K}+k}$ using Algorithm 1 and receive score $\mathcal{R}_{n-1}^{\mathcal{K}+k}$.
        
        Analyze error-prone points of questions in $\mathcal{C}_{n-1}^{\mathcal{K}+k}$ using Algorithm 3 and receive new $\mathcal{C}_{n-1}^{\mathcal{K}+k}$.
        
        Set $\mathcal{D} \cup (\mathcal{C}_{n-1}^{\mathcal{K}+k}, \mathcal{R}_{n-1}^{\mathcal{K}+k}) \to \mathcal{D}$.
    }
    D.append(D').
    
    Sort D by scores in ascending order.
    
    Set D[: max length] $\to$ D.
    
    Set knowledge background $\Vert \mathcal{D} \Vert$ optimization task $\to \mathcal{P}_n$.
}
return $\mathcal{D}$.
\end{algorithm}

\subsection{\textbf{Instructional Design Analysis}}
\subsubsection{\textbf{Analyst Agent}}
The Analyst Agent explains the common mistakes produced by questions in the instructional design to alert students. We compile 50 common mistakes in algebraic equations into five groups. For each example from Alg.~\ref{Alg2}, the Analyst Agent identifies three common error-prone points for students with corresponding knowledge backgrounds. These error points are inserted into the example explanations to help students avoid these errors.

\subsubsection{\textbf{Error-prone Point Analysis Algorithm}}
To identify common errors students may make, we develop the Analyst Agent Expert Analysis (AAEA) algorithm. The algorithm extracts examples from instructional designs. Utilizing annotated common errors as reference, the Analyst Agent generates potential errors that students might make for each example. These errors are inserted into the example explanations within the instructional design, serving as reminders to help students avoid these mistakes and correctly solve the problems. Alg.~\ref{Alg3} provides a detailed description of this process.

\begin{algorithm}[t]
\label{Alg3}
\small
\caption{Analyst Agent Expert Analysis (AAEA)}

\KwIn{Question Analyzer $A_a$, Common Mistakes Database ($CMD$), Instructional Design ($Lp$), Example Questions ($EQs$) generated by Optimizer Agent, Knowledge Background ($KB$)}

\KwOut{new instructional designs with integrated error-prone points and probabilities($nlp$)}

$A_a$ $\leftarrow$ Initialize Analyst Agent

$EQs$ $\leftarrow$ Retrieve example questions from $Lp$

Load $CMD$

\For{$eq$ in $EQs$} {

    $KB$ $\leftarrow$ Identify target student group knowledge background for $eq$
    
    $CMs$ $\leftarrow$ Retrieve top 3 common mistakes from $CMD$ based on $KB$ and mistake probability

    \For{$cm$ in $CMs$} {
    
        $mp$ $\leftarrow$ Calculate mistake probability for $eq$
        
        $me$ $\leftarrow$ Generate explanation for $cm$
    }
    $CMs$ $\leftarrow$ Sort $CMs$ by $mp$ in descending order
    
    Insert $CMs$ and $me$ into $eq$ section of $Lp$
    
}
\textbf{return} $nlp$
\end{algorithm}

\section{\textbf{EXPERIMENTS}}
In this section, we verify the results of the generation framework. We propose an instructional design evaluation system (CIDPP) to analyze the quality of instrutional designs. In addition, we conducted ablation experiments to demonstrate the roles of each module in the framework.

\begin{figure*}[!htbp]
\centering
\includegraphics[width=0.8\textwidth]{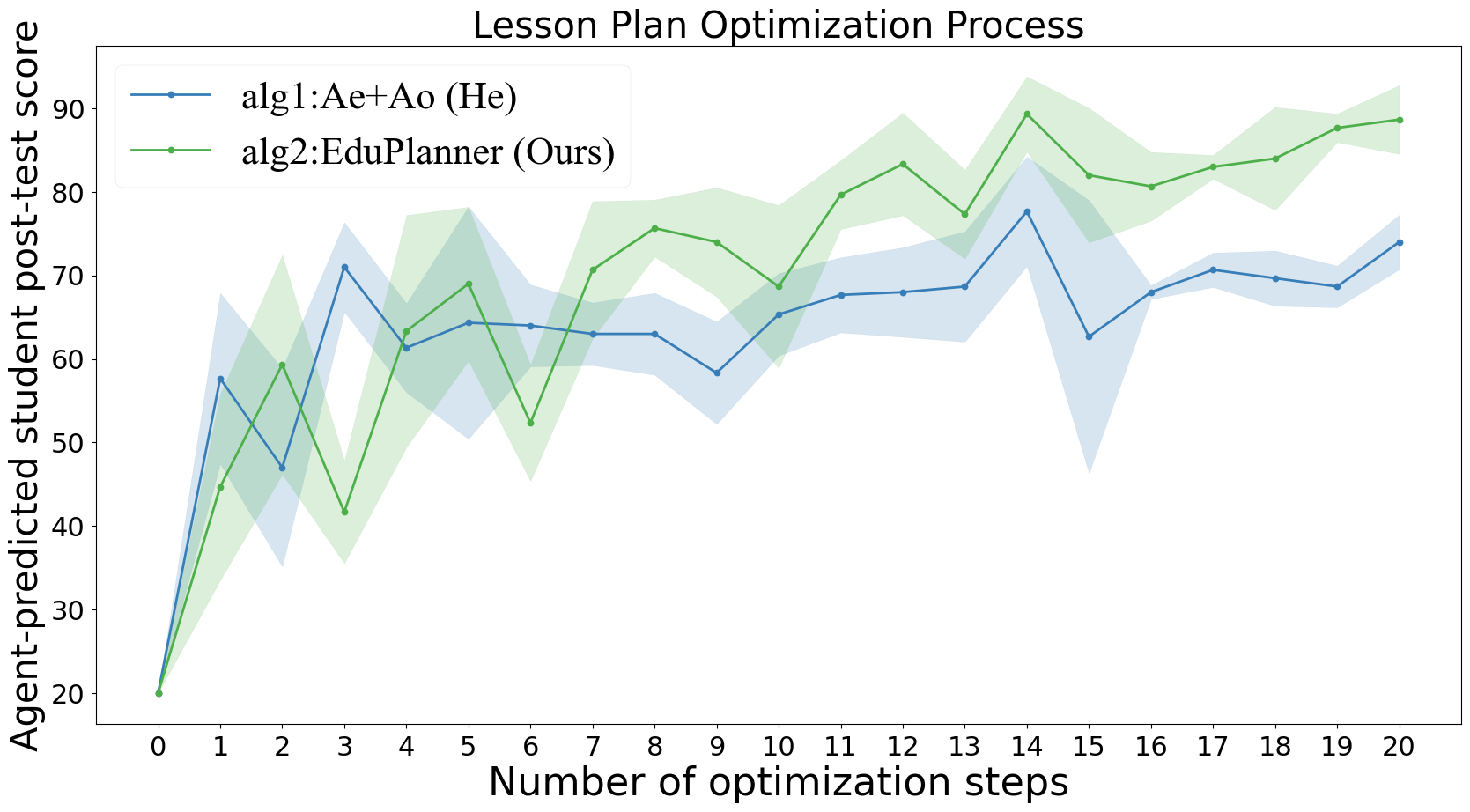}\\
\caption{In this figure, we obtained the optimization process curves of our framework and He's method~\cite{he2024evaluating} under the premise that the hyperparameters are consistent with the initial instructional design for curriculum and learning activities. Our framework has a higher score in the instructional design than He's~\cite{he2024evaluating} method, and the optimization process is smoother.}
\label{Contrastive Results}
\end{figure*}


\subsection{\textbf{Hyperparameter Configuration}}

In this section, we tune the hyperparameters of each component to ensure optimal performance. For the evaluator agent's base model, \emph{i.e.} \texttt{Meta-Llama-3-70B-Instruct}, we set the temperature parameter at $0.0$ for stable output. The optimizer agent's base model, \emph{i.e.} \texttt{GPT-4}, has a temperature parameter set at $1.0$ for diverse instructional design content. The analyzer, also based on \texttt{GPT-4}, has a temperature parameter set at $0.7$ for balance between stability and diversity.

\subsection{\textbf{Contrastive Results}}
To validate the efficacy of our framework, we compare our results with other methods by LLMs like \texttt{GPT-3.5-turbo}, \texttt{Llama-3-70B-Instruct} and \texttt{GPT-4} as detailed in the Fig~\ref{Contrastive Results}. For the direct generation of instructional designs through LLMs, we apply a prompt as below ``\textit{Generate an instructional design with the teaching theme of algebraic equations. It contains two parts, the explanation of knowledge points and the explanation of examples. The explanation of knowledge points should be comprehensive, and the explanation of examples should be representative. The generated instructional designs must be concise and limited to $200$ words.}'' Finally, we assess the instructional design based on the CIDPP as shown in Tab.~\ref{Indicators Results}, where the results objectively demonstrate the effectiveness of our framework in improving the instructional designs. The prompt of evaluation is as follow: 

``\textit{\textbf{Role}: You are an impartial evaluator, experienced in educational content analysis and instructional design evaluation.
\textbf{Attention}: You are responsible for assessing the quality of a given instructional design based on five specific evaluation criteria. Your evaluation should be objective and based solely on the Evaluation Standard provided below. \textbf{Lesson Plan}: $\{$lessonplan$\}$}
\textit{\textbf{Evaluation Standard}: 
Clarity: 
The lesson plan's directness and simplicity, ensuring it avoids unnecessary complexity and redundancy.
Integrity:
Whether the lesson plan is complete and systematic, covering both knowledge point explanations and exercise explanations in a complementary manner.
Depth:
The ability of the lesson plan to inspire deep thinking and facilitate understanding of the underlying connections between knowledge points.
Practicality:
The practical application value of the examples in the lesson plan, ensuring students can use the knowledge to solve real-life problems.
Pertinence:
The adaptability of the lesson plan to different students' knowledge levels and learning needs to achieve optimal learning outcomes.
\textbf{Constraints}:
Avoid any bias in evaluation based on the content's length or appearance.
Be as objective as possible in assessing each aspect individually without favoring any specific structure or terminology.
\textbf{Workflow}:
Output your final verdict in the following format:"[A]: [points]; [short analyzes]", "[B]: [points]; [short analyzes]", "[C]: [points]; [short analyzes]", "[D]: [points]; [short analyzes]", "[E]: [points]; [short analyzes]". Take a deep breath and think step by step!''}

In our research, we introduce a novel framework designed to enhance the optimization of instructional design for curriculum and learning activities. Based on baseline~\cite{he2024evaluating}, we preserve the evaluator and optimizer agent, while integrating two crucial additions: a Skill-Tree structure and an analyst agent. As illustrated in Fig.~\ref{Contrastive Results}, our method outperforms baseline~\cite{he2024evaluating} and exhibits a smoother process.

\begin{figure*}[!htbp]
\centering
\includegraphics[width=0.8\textwidth]{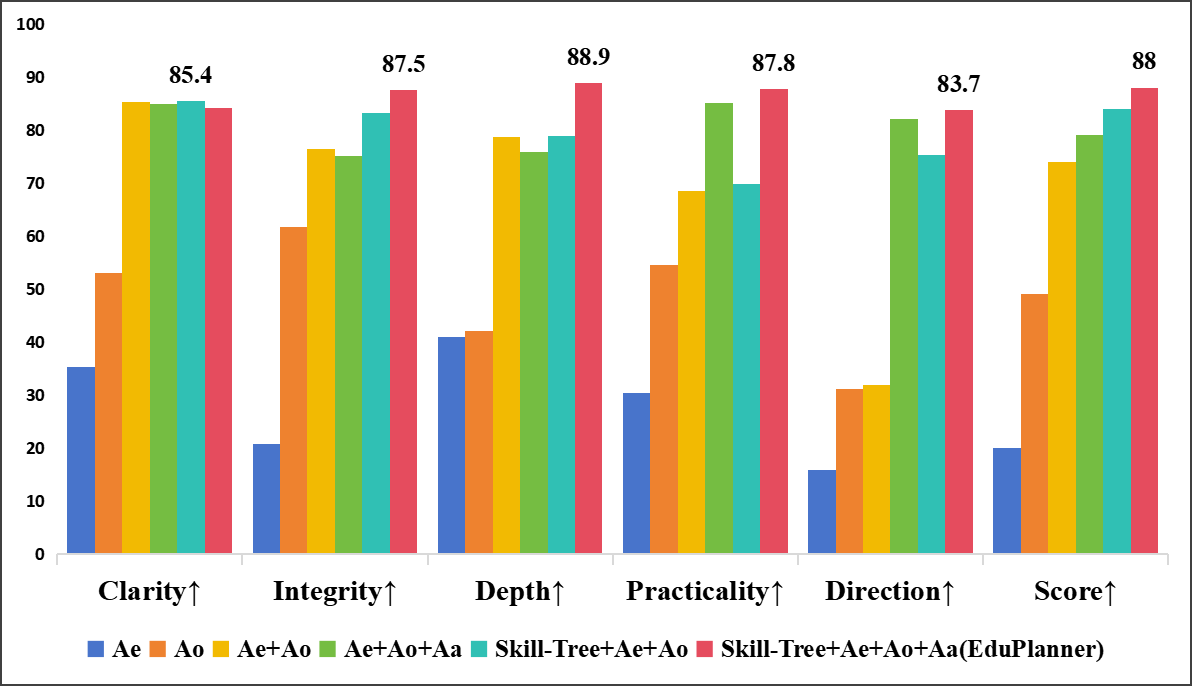}
\caption{The figure illustrates the influence of each framework component on each evaluation metric in CIDDP via a bar chart. }
\label{6666}
\end{figure*}

\begin{figure}[t]
\centering
\includegraphics[width=\linewidth]{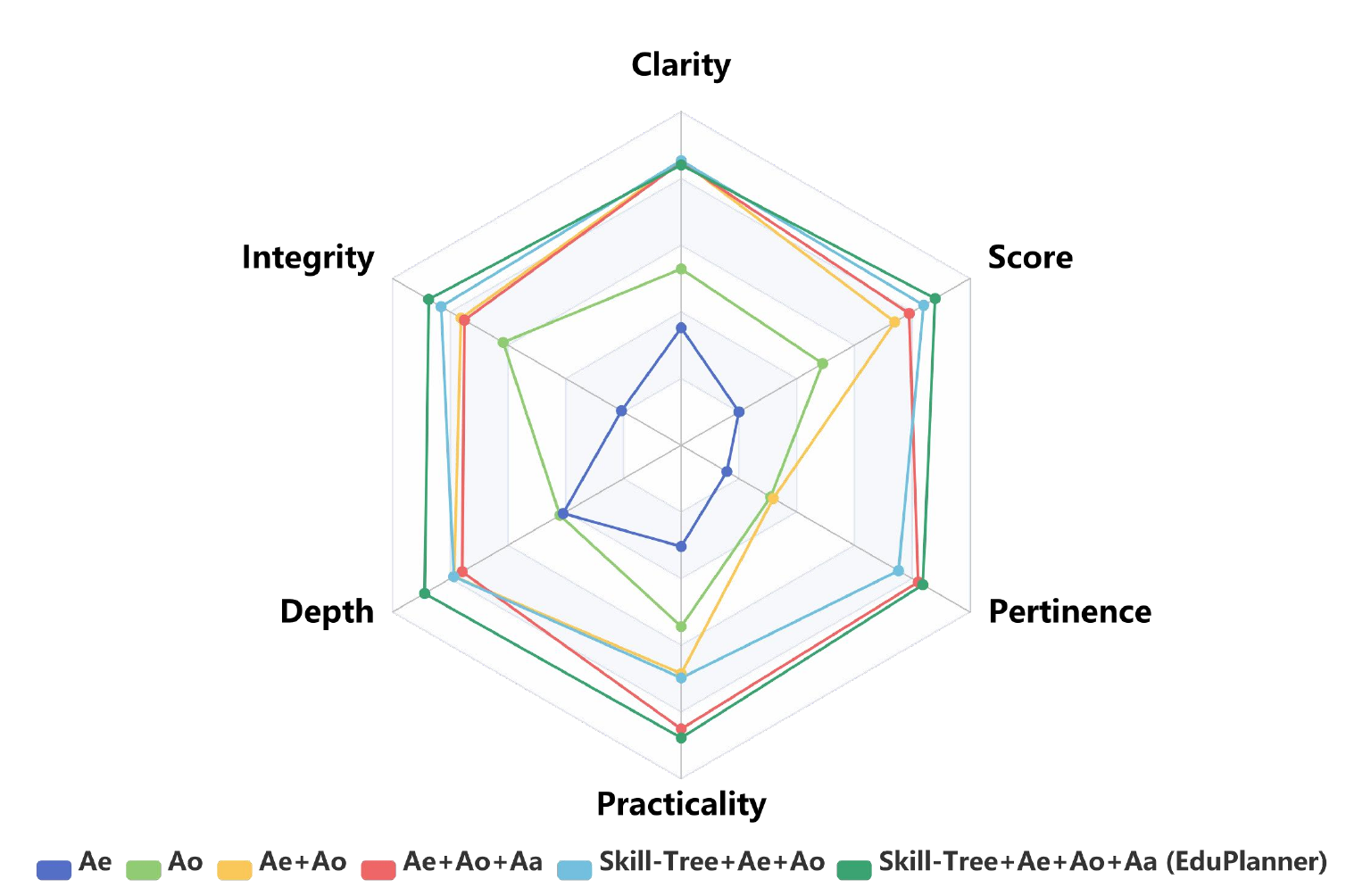}
\caption{The figure employs a radar chart to portray the incremental effects of each component.}
\label{radar}
\end{figure}

Furthermore, we conduct an ablation study on all components of our instructional design in an optimization framework. As shown in Tab.~\ref{table3} and Fig.~\ref{6666} and Fig.~\ref{radar}, we evaluate the individual impact of each component and measure its contribution to the optimization results using CIDPP.

These contrastive results emphasize that the integration of the evaluator agent, optimizer agent, analyst agent, Skill-Tree mechanisms, and CIDPP within our methodology improves the structural integrity of instructional design and enhances their quality and pedagogical resonance through personalized improvements. Our experiments have shown that utilizing a standalone evaluator agent or an optimizer agent does not produce the most favorable results. But integrating these agents establishes a great baseline, excelling in the dimensions of clarity, integrity, and depth. Moreover, the addition of an analyst agent has notably improved the practicality and pertinence of instructional design for curriculum and learning activities.
Although the enhancement in clarity may not be pronounced, the Skill-Tree mechanism improves other performance metrics by creating a map of the student's competencies and crafting educational content that is aligned with their individual knowledge background.

\section{\textbf{DISCUSSION}}
The advent of AGI has ushered a transformation across diverse sectors, presenting both opportunities and challenges for the field of education. It extends novel forms of support for teaching and learning, introducing innovative methodologies to educational research with a focus on multidisciplinary integrated approaches like pedagogical~\cite{ciolacu2021education} and AI. 
Smart education~\cite{yang2020using,tham2023smart,wan2023pedagogical,singh2020smart}, identified as a pivotal trend in the future educational landscape, is obtaining increasing attention and importance. Central to this trend is personalized education, which disrupts traditional teacher-student roles by dynamically adjusting educational resources and teaching strategies to meet individual students' needs. 

Developing instructional design for curriculum and learning activities is a crucial aspect of educational practice that must adapt to changing requirements. The traditional method of creating instructional designs is time-consuming for teachers, which involves understanding each student's individual characteristics and knowledge level to effectively align subject matter with their needs.
As educational concepts evolve, previous teaching methods may no longer meet the needs of current students.
Developing effective instructional design requires the integration of clear assessment methods and feedback mechanisms for quick adjustments and improvements in teaching strategies, which still presents a challenge. 
Modern instructional design often involves using educational technologies like multimedia tools and online platforms. Educators face the added challenge of mastering these tools and integrating them into their teaching practices.

Considering the existing issues, our framework introduces an evaluative system for instructional designs, a Skill-Tree model for student ability, and a structure with three collaborative agents. These advancements facilitate a comprehensive evaluation and improvement of instructional design quality, customized to address individual student needs.

In our studies, we propose a novel evaluation system for instructional designs that considers clarity, integrity, depth, practicality, and pertinence. This evaluation system reflects the pedagogical principles in instructional design. 
The instructional design's clarity helps teachers establish clear teaching objectives, ensuring that teaching activities are purposeful and in line with goal-oriented teaching principles in pedagogical philosophy. 
Ensuring the integrity of instructional designs emphasizes essential elements such as detailed explanations of key concepts and relevant examples, aligning with the comprehensive learning principles promoted in educational philosophy.
The depth of instructional designs measures how well the instructional design engages students in deep thinking and provides detailed explanations of the deep interrelations of knowledge, aligning with the pedagogical emphasis on deep-level learning.
The practicality of instructional designs evaluates how effectively students can apply their knowledge to solve real-world problems. It corresponds with the pedagogical philosophy that emphasizes the application of knowledge to address practical issues.
The pertinence focuses on tailoring instructional designs to meet students' varying knowledge levels and learning requirements, aligning with the personalized teaching philosophy promoted by pedagogical philosophy. 

\begin{table*}[htbp]
\centering
\small
\caption{Quality Indicators}
\label{Indicators Results}
\begin{tabular}{c|ccccc|c}
\toprule
\textbf{Strategy} & \textbf{Clarity↑}& \textbf{Integrity↑}& \textbf{Depth↑}& \textbf{Practicality↑}& \textbf{Pertinence↑}& \textbf{Score↑}\\ \hline
GPT-3.5-turbo & 28.2 & 28.1 & 33.6 & 33.8 & 10.0 & 17 \\ 
Llama-3-70B-Instruct & 35.2 & 20.7 & 40.9 & 30.4 & 15.8& 20 \\ 
GPT-4 & 52.9 & 61.7 & 42.0 & 54.4 & 31.0 & 49 \\ \
EduPlanner & \textbf{84.1} & \textbf{87.5} & \textbf{88.9} & \textbf{87.8} & \textbf{83.7} & \textbf{88} \\ \hline
\end{tabular}

\end{table*}

\begin{table*}[htbp]
\centering
\small
\caption{Ablation study on the Gsm8k and Algebra datasets.}
\label{table3}
\begin{tabular}{c|ccccc|c}
\toprule

\textbf{Strategy} & \textbf{Clarity↑} & \textbf{Integrity↑} & \textbf{Depth↑} & \textbf{Practicality↑} & \textbf{Pertinence↑} & \textbf{Score↑} \\ \hline
Ae & 35.2 & 20.7 & 40.9 & 30.4 & 15.8 & 20 \\ 
Ao & 52.9 & 61.7 & 42.0 & 54.4 & 31.0& 49 \\ 
Ae+Ao (He~\cite{he2024evaluating}) & 85.3 & 76.3 & 78.7 & 68.5 & 31.9 & 74 \\ 
Ae+Ao+Aa & 84.9 & 75.1 & 75.9 & 85.1 & 82.1 & 79 \\ 
Skill-Tree+Ae+Ao & \textbf{85.4} & 83.2 & 78.8 & 69.8 & 75.3 & 84 \\
Skill-Tree+Ae+Ao+Aa (EduPlanner) & 84.1 & \textbf{87.5} & \textbf{88.9} & \textbf{87.8} & \textbf{83.7} & \textbf{88} \\
\hline
\end{tabular}
\end{table*}

\section{\textbf{CONCLUSION}}
In this paper, we develop EduPlanner, a multi-agent system comprising three key intelligent agents: the Evaluator, the Optimizer, and the Example Analyzer, for the automated evaluation and optimization of instructional designs. To meet students with varying knowledge levels, we design a Skill-Tree structure to measure their knowledge bases, which allows the system to capture the differences in each student's learning abilities and generate corresponding instructional designs. Through a series of experiments, we validate our framework's superior performance over other methods. We also conducted ablation experiments to examine the roles of the Skill-Tree structure and the Analyst Agent. The results confirm that these novel components are crucial in generating remarkable instructional designs that show the transformative potential of AGI to address complex problems and complex challenges in the domain of education.

\section{\textbf{LIMITATIONS AND FUTURE WORK}}
\subsection{\textbf{Limitations}}
While EduPlanner has achieved notable results in the generation and improvement of instructional design, it faces various limitations due to the constantly changing educational technology environment and the increasing diversity of teaching needs. 

1) The skill tree is specifically tailored to generate instructional designs within certain subjects, which may limit its applicability to other disciplines.

2) EduPlanner's reliance on LLM resources is marked by substantial consumption, which leads to a huge cost in terms of time and money. 

3) There is an absence of adaptive integration with existing tools within practical teaching environments, which indicates a demand for ensuring alignment with current educational practices.

\subsection{\textbf{Future Work}}
GAI is revolutionizing instructional designs evaluation and optimization, aiming to better address the evolving needs of educational systems~\cite{hung2019improving,zhu2023stable,salem2019knowledge,wangoo2020smart,sutjarittham2019experiences}. Despite current challenges, continuous improvements in these systems are expected to boost the educational content. Future research should focus on creating models tailored to specific subjects and improving LLMs to better replicate real-world educational settings. Another key strategy involves the combination of various disciplines, improving students' skills in integrating knowledge across different subjects.
The knowledge developed through optimizing instructional designs influences other smart education domains. Incorporating LLMs into instructional designs can enhance the adaptability and effectiveness of educational practices in response to the dynamic educational environment.

\bibliography{ref} 
\end{document}